\documentclass{article}
\usepackage{arxiv}
\usepackage{url}            
\usepackage[utf8]{inputenc}
\usepackage{amssymb,amsthm}
\usepackage{enumitem}
\usepackage{microtype}
\usepackage{idxlayout}
\usepackage{lmodern}
\usepackage{float}
\usepackage{amsthm}
\usepackage{graphicx}
\usepackage{mathtools}
\usepackage{longtable,booktabs}
\usepackage{subcaption}
\usepackage{fullpage}
\usepackage{times}
\usepackage[rightcaption]{sidecap}
\sidecaptionvpos{figure}{t}
\usepackage{fancyhdr,graphicx,amsmath,amssymb}
\usepackage{wrapfig,lipsum,booktabs}
\usepackage[ruled,vlined, linesnumbered]{algorithm2e}
\include{pythonlisting}

\makeatletter
\newcommand{\printfnsymbol}[1]{%
  \textsuperscript{\@fnsymbol{#1}}%
}
\makeatother

\title{Soft Q network}

\author{
  Jingbin Liu \\
  CreateAmind\\
  \And
 Shuai Liu\\
  CreateAmind\\
     \And
 Xinyang Gu \\
  CreateAmind\\
}

\begin{document}
\maketitle

\begin{abstract}

Deep Q Network (DQN) is a very successful algorithm, yet the inherent problem of reinforcement learning, i.e. the exploit-explore balance, remains. In this work, we introduce entropy regularization into DQN and propose SQN. We find that the backup equation of soft Q learning can enjoy the corrective feedback if we view the soft backup as policy improvement in the form of Q, instead of policy evaluation. We show that Soft Q Learning with Corrective Feedback (SQL-CF) underlies the on-plicy nature of SQL and the equivalence of SQL and Soft Policy Gradient (SPG). With these insights, we propose an on-policy version of deep Q learning algorithm, i.e. Q On-Policy (QOP). We experiment with QOP on a self-play environment called Google Research Football (GRF). The QOP algorithm exhibits great stability and efficiency in training GRF agents.

\end{abstract}

\keywords{RL \and ENTROPY \and SQN \and EFFICIENCY}

\section{INTRODUCTION}\label{header-n1}
When DQN is announced by DeepMind in 2015, the whole world is surprised by its simplicity and capability to solve all kinds of Atari games \cite{DQN}. The DeepMind system uses a deep convolutional neural network to represent the Q function, and experience replay buffer is introduced to improve training stability by removing correlations in the observation sequence and smoothing changes in the data distribution. Since then, many improvements have been made to DQN. Besides the algorithmical modification, distributed training frameworks are also developed so as to solve large scale environments.
Double DQN \cite{DDQN} addresses the overestimation problem of Q-function by using two Q networks to decouple action selection and Q-value evaluation. Prioritized experience replay \cite{PER} improves data efficiency by replaying more with transition tuples that is less fitted. Ape-X DQN proposes a distributed architecture for Q-learning at scale by decoupling acting from learning \cite{APEX}. Recurrent Replay Distributed DQN (R2D2) investigates the training of RNN-based RL agents from distributed prioritized experience replay  \cite{R2D2}.

In spite of the great improvements made on DQN, the inherent problems of DQN remain. 
First of all, it can suffer from the sub-optimum problem. In the algorithm design, a decaying exploration factor needs to be artificially added and tuned in order to let learning happen. It usually results in a solution doing well in the short term, but under-performing in the long run.  Secondly, like many other reinforcement learning methods, DQN tends to have low sample-efficiency. Take the standard baseline environments Atari57 for example, billions of interaction steps are needed to solve each one of the environments\cite{R2D2}\cite{APEX}. Last but not least, DQN is very sensitive with respect to its hyper-parameters. Parameter tuning can be time-consuming and harm reproducibility. Due to these shortcomings, large scale reinforcement learning projects like DOTA2 and StarCraftII \cite{Dota2}\cite{AlphaStar} use policy gradient based algorithms instead of algorithms based on Q-learning.

In reinforcement learning, explore-exploit trade-off is an eternal problem. In order to balance the explore-exploit trade-off, we introduce entropy regularization into DQN and propose Soft Q Network (SQN) in this work. Entropy regularization is an important idea in reinforcement learning \cite{SQL}. In addition to the environment reward, entropy regularization brings in an entropy reward which encourages exploration. SQN no longer needs the intractable exploration factor, instead the exploration of SQN is controlled by an adaptive parameter $\alpha$ called temperature.

Q-value based deep reinforcement learning algorithms apply approximate dynamic programming (ADP) with function approximation. ADP methods rely on previous Q-functions to generate targets for training the current Q-function. As pointed out in \cite{DisCor}, because the target values are computed by applying a Bellman backup on the previous Q-function, $\Hat{Q}$ (target value), rather than the optimal $Q^{*}$, errors in $\Hat{Q}$ at the next states can result in incorrect Q-value targets at the current state. No matter how often the current transition is observed, or how accurately Bellman errors are minimized, the error in the Q-value with respect to the optimal Q-function $|Q {-} Q^{*}|$ at this state is not reduced. We show that "the corrective feedback problem" can be eliminated theoretically for Soft Q-Learning (SQL) with a small modification on the standard SQL algorithms. We call the new SQL algorithm variant Soft Q-Learning with Corrective Feedback (SQL-CF), which includes Soft Q Network with Corrective Feedback (SQN-CF) and Soft Actor-Critic with Corrective Feedback (SAC-CF) for the discrete and continuous cases, respectively.

In this work, we also propose the Q On-Polcy (QOP) algorithm, a distributed variant of SQN-CF with n-step backup. It is worth noting that SQN is an off-policy algorithm while QOP is an on-policy algorithm. It can be proven that SQL equals Soft Policy Gradient (SPG) \cite{schulman2017equivalence} \cite{SPG}. Since policy gradient is an on-policy algorithm, we speculate that SQL can also been trained in an on-policy fashion. In order to verify this extrapolation, we experiment with the QOP algorithm on an self-play reinforcement learning environment called Google Research Football (GRF), a new reinforcement learning environment where agents are trained to play football \cite{GRF}. We identify the strong correlation between the reuse ratio of sampling data and training speed, which indicates the on-policy nature of the QOP algorithm. Our experiment shows the training stability and efficiency of SQN and QOP, which can generate strong agents with diverse attacking and defending strategies.

\section{BACKGROUND}\label{header-n2}

\subsection{MDP}\label{header-n21}

The key problem of RL is searching for a policy that maximizes the accumulate future rewards 
in a Markov Decision Process \((\text{MDP})\) defined by
the tuple \((\mathcal{S,A,P,R})\) \cite{planet}. 

\begin{itemize}
\item
  \(\mathcal {S}\) represents a set of states
\item
  \(\mathcal {A}\) represents a set of actions, 
\item
  \(\mathcal {P:S\times A\to P(S)}\) stands for the transition function which maps
  state-actions to probability distributions over next states
  \(\text{P}(s^{\prime}|s,a)\)
\item
  \(\mathcal {R}:\mathcal{S\times A }\to \mathbb{R} \) corresponds to the
  reward function, with \(r_t=R(s_t,a_t)\)
\end{itemize}

Within this framework, the agent acts in the environment according to a policy $a  \sim \pi(\cdot|s)$.
the environment changes to a new state following $s^{\prime} \sim \mathcal{P}(\cdot|s, a)$ and provides a reward $r$. The agent iterates its policy based on the information gathered via interacting with the environment to obtain an optimal policy:
\begin{equation}
    \pi^* = \arg \underset{\pi}\max \E{\tau \sim \pi}{ \sum_{t=0}^{\infty} \gamma^t  r(s_t, a_t) }   
\end{equation}{}

\subsection{Maximum Entropy Reinforcement Learning}\label{header-n22}

The maximum entropy reinforcement learning generalizes the standard objective with an entropy term, such that the optimal policy additionally aims to maximize its entropy at each visited state \cite{SQL}. The temperature parameter $\alpha$ that determines the relative importance of the entropy term versus the reward is account for the explore-exploit balance.
\begin{equation}
    \pi^* = \arg \underset{\pi}\max\E{\tau \sim \pi}{ \sum_{t=0}^{\infty} \gamma^t \bigg( r(s_t, a_t) + \alpha H\left(\pi(\cdot|s_{t+1})\right) \bigg)}.
\end{equation}{}

The temperature parameter $\alpha$ can be tuned to encourage exploration. When $\alpha$ is large, the policy will be more stochastic, on the contrary, when $\alpha$ is small, the policy will become more deterministic. In the limit $\alpha \rightarrow 0$, We can recover exactly the standard reinforcement learning. The maximum entropy objective has a number of conceptual and practical advantages. The policy is incentivized to explore more widely, capturing multiple modes of near-optimal behavior while giving up on unpromising avenues. Unlike the standard reinforcement learning, in problem settings where multiple actions seem equally attractive, the policy will commit equal probability mass to those actions instead of collapsing into one action randomly.

In the entropy-regularized framework, \(Q^{\pi}\) should be modified to include the entropy term: 
\begin{align}
    Q^{\pi}(s,a) &= \E{\tau \sim \pi}{ \left. \sum_{t=0}^{\infty} \gamma^t \bigg( r(s_t, a_t) + \alpha  H\left(\pi(\cdot|s_{t+1})\right) \bigg) \right| s_0 = s, a_0 = a}.
    \label{eq:q}
\end{align}
We can define the value function \(V^{\pi}\) as:
\begin{align}
    V^{\pi}(s) &= \E{a \sim \pi}{Q^{\pi}(s,a)+ \alpha  H\left(\pi(\cdot|s)\right)},
\end{align}
and rewrite the modified Bellman backup of \(Q^{\pi}\) as:
\begin{align}
    Q^{\pi}(s,a) &= r(s,a)  + \gamma \E{s' \sim P}{V^{\pi}(s')}. \label{eq:backup}
\end{align}

\section{METHOD}\label{header-n3}

\subsection{Soft Q Network}
We will begin by introducing the soft policy evaluation step of soft policy iteration. The soft Q-value for a fixed policy with  the maximum entropy objective can be computed iteratively, following the Bellman backup according to Eq.(\ref{eq:backup}).
With the modified Bellman backup operator $\mathcal{T}^\policy$, we have:
\begin{align}
\mathcal{T}^\policy Q(\st, \at) \triangleq  \reward(\st, \at) + \discount \E{\stp \sim \pdyn}{V(\stp)}, \label{eq:soft_bellman_backup_op} 
\end{align}
where
\begin{align}
V(\st) = \E{\at\sim\policy}{\Q(\st, \at) - \alpha\log\policy(\at|\st)}
\label{eq:soft_value_function}
\end{align}
is the soft state value function. We can obtain the soft Q-function for any policy $\policy$ by repeatedly applying $\Q^{k+1} = \mathcal{T}^\policy \Q^k$ given an arbitrary starting $Q^0$. The sequence $Q^k$ will converge to the soft Q-function of $\policy$ as $k\rightarrow \infty$.

For discrete reinforcement learning, it is convenient to define the policy from the learned Q function without using an additional policy network, such as DQN. Different from DQN, Soft Q Network uses a softmax policy instead of a hard max policy. We define the policy as:
\begin{align}
\policy(\voidarg|\st) &= \text{softmax}  \left(  Q (\st, \voidarg) / \alpha \right) \notag\\
& = {\exp\left(Q (\st,\voidarg) / \alpha  - \log Z (\st)\right)}
\label{eq:softmax policy}
\end{align}
where $Z (\st)$ is the partition function for normalization:
\begin{align}
Z (\st) = \sum_{i=1}^n \exp\left(Q (\st, a_t^i)/ \alpha \right)
\label{eq:z}
\end{align}
It can be proven that Eq.(\ref{eq:softmax policy}) is a policy improvement, i.e. the soft policy improvement.
Let $\Q^\policyold$ and $\V^\policyold$ be the corresponding soft state-action value and soft state value of policy $\policyold$, and let $\policynew$ be defined according to Eq.(\ref{eq:softmax policy}) as:
\begin{align}
\policynew(\voidarg|\st) &= {\exp\left(Q^\policyold (\st,\voidarg) / \alpha  - \log Z^\policyold (\st)\right)}. \label{eq:pi_new}
\end{align}
By using the non-negative property of Kullback–Leibler divergence
\begin{align}
\kl{\policyold(\voidarg|\st)}{\policynew(\voidarg|\st)} \geq \kl{\policynew(\voidarg|\st)}{\policynew(\voidarg|\st)} =0
\end{align}
and substituting $\policynew$ on the right-hand side of Kullback–Leibler divergence with Eq.(\ref{eq:pi_new}), the inequality can be expanded as:
\begin{align}
\resizebox{1\textwidth}{!}{$
\E{\at\sim\policyold}{\log \policyold(\at|\st) - Q^\policyold(\st,\at)/\alpha + \log Z^\policyold(\st)}
\geq  \E{\at\sim\policynew}{\log \policynew(\at|\st) - Q^\policyold(\st, \at)/\alpha + \log Z^\policyold(\st)}
$}.
\end{align}
Since partition function $Z^\policyold$ depends only on the state, we have:
\begin{align}
 V^\policyold(\st) \leq \E{\at\sim\policynew}{Q^\policyold(\st, \at) - \alpha \log \policynew(\at|\st)} .
\label{eq:soft_value_bound}
\end{align}
By applying the soft Bellman equation iteratively, we can confirm the soft policy improvement \cite{SQL} \cite{SAC1}.
\begin{align}
Q^\policyold(\st, \at) &= \reward(\st, \at) + \discount\E{\stp\sim\pdyn}{V^\policyold(\stp)}  \label{eq:bell1}\\
&\leq \reward(\st, \at) + \discount\E{\stp\sim\pdyn}{\E{\atp\sim\policynew}{Q^\policyold(\stp, \atp) - \alpha \log \policynew(\atp|\stp)}}  \label{eq:bell2} \\
&\ \  \vdots\notag\\
& \leq Q^\policynew(\st, \at). \label{eq:bell3}
\end{align}

\begin{figure}
    \centering
    \includegraphics[width=0.795\textwidth]{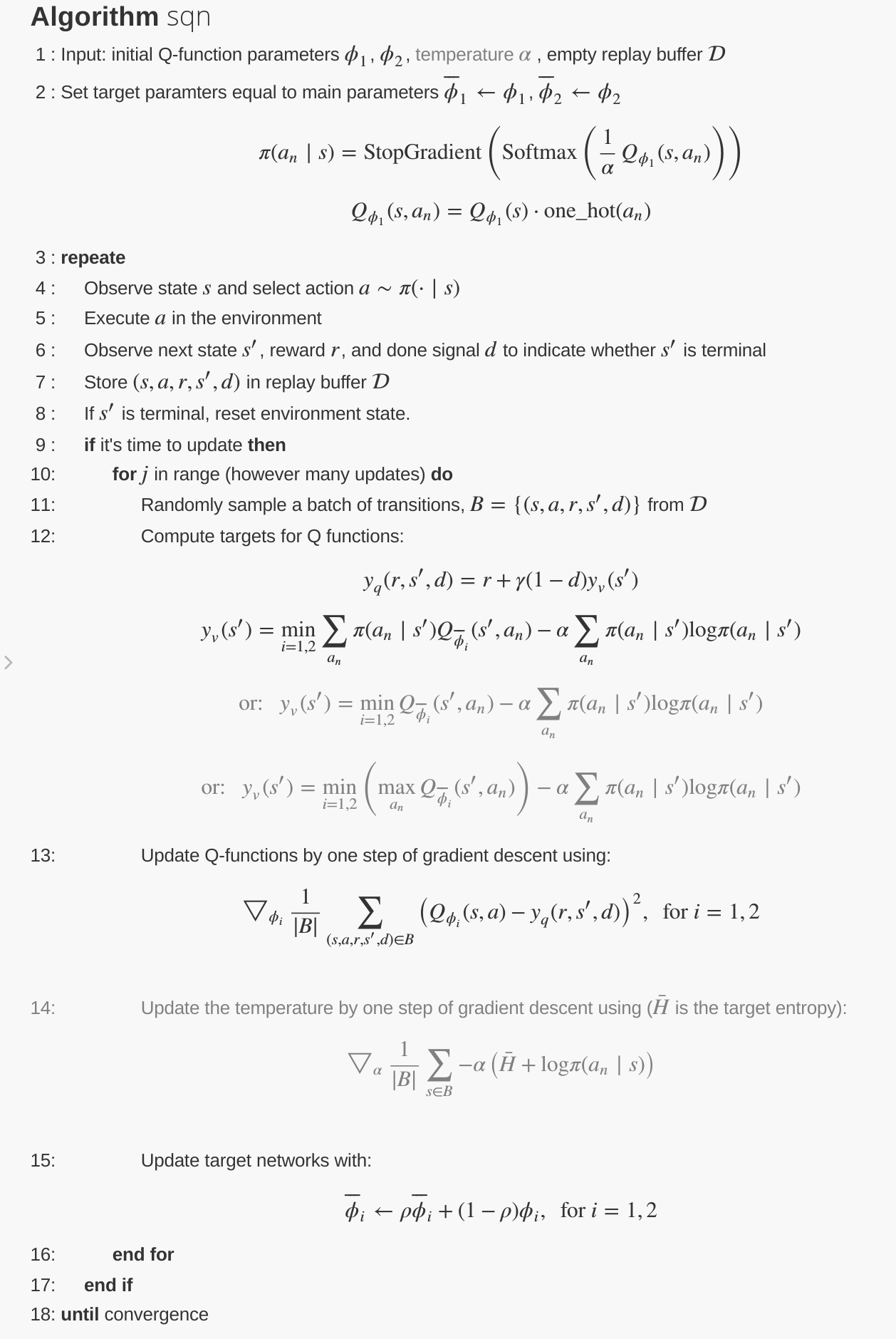}
    \caption{The SQN algorithm. There are two Q function networks in the framework. Q function networks are updated with Q value backups and the policy is defined as the softmax of the learned Q function.}
    \label{fig:sqn}
\end{figure}

Combining the soft policy evaluation and the soft policy improvement, we have the soft policy iteration of SQN. The double Q-learning trick is applied in SQN to mitigate over-estimation of the Q functions \cite{TD3}. In addition to the double Q networks, two more target Q networks are introduced to stabilize training. The two target Q networks are updated by polyak averaging the parameters of the two main Q networks over the course of training. The pseudo-code of SQN is presented in Fig.\ref{fig:sqn}, where we propose three different schemes for the backup of soft Q functions.

\subsection{SQN with Corrective Feedback}

In the policy evaluation step, Bellman backup relies on previous Q-functions to generate targets for training the current Q-function. As shown in \cite{DisCor}, deep Q learning can have "the corrective feedback problem". If the target values are computed by applying a Bellman backup on the optimal $Q^{*}$, the Q-function is updated with corrective feedback. But in practice the Q-function is updated with non-corrective Q-value targets, which can harm training efficiency.

Here we show that SQL can be modified to enjoy the corrective feedback. Instead of viewing the Bellman backup equation as policy evaluation, the soft Bellman backup equation can be transformed to be policy improvement in the form of Q value. Let $\policynew$ be defined as Eq.(\ref{eq:pi_new}):
\begin{align}
\policynew(\voidarg|\st) &= {\exp\left(Q^\policyold (\st,\voidarg) / \alpha  - \log Z^\policyold (\st)\right)}, \label{eq:pi_new1}
\end{align}
and define a new soft Bellman backup of Q function:
\begin{align}
Q(\st, \at) = \reward(\st, \at) + \discount\E{\stp\sim\pdyn}{\E{\atp\sim\policynew}{Q^\policyold(\stp, \atp) - \alpha \log \policynew(\atp|\stp)}}, \label{eq:backup1}
\end{align}
from Eq.(\ref{eq:bell2}), we have:
\begin{align}
Q(\st, \at) \geq  Q^\policyold(\st, \at). \label{eq:backup2}
\end{align}
Substituting $Q^\policyold(\st, \at)$ in Eq.(\ref{eq:pi_new1}) with $Q(\st, \at)$ in Eq.(\ref{eq:backup1}), i.e. $\policy(\voidarg|\st) = {\exp\left(Q (\st,\voidarg) / \alpha  - \log Z (\st)\right)}$ and applying the above inequality and the soft policy improvement inequality, we obtain:
\begin{align}
Q^\policy(\st, \at) \geq  Q(\st, \at) \geq Q^\policyold(\st, \at). \label{eq:policyimprove}
\end{align}
Repeating the two steps of Eq.(\ref{eq:pi_new1},\ref{eq:backup1}), we obtain the policy improvement in the form of $Q$ value according to Eq.(\ref{eq:policyimprove}). In other words, every Bellman backup step of SQN-CF is guaranteed to be a policy improvement and has corrective feedback.

\begin{figure}
    \centering
    \includegraphics[width=0.795\textwidth]{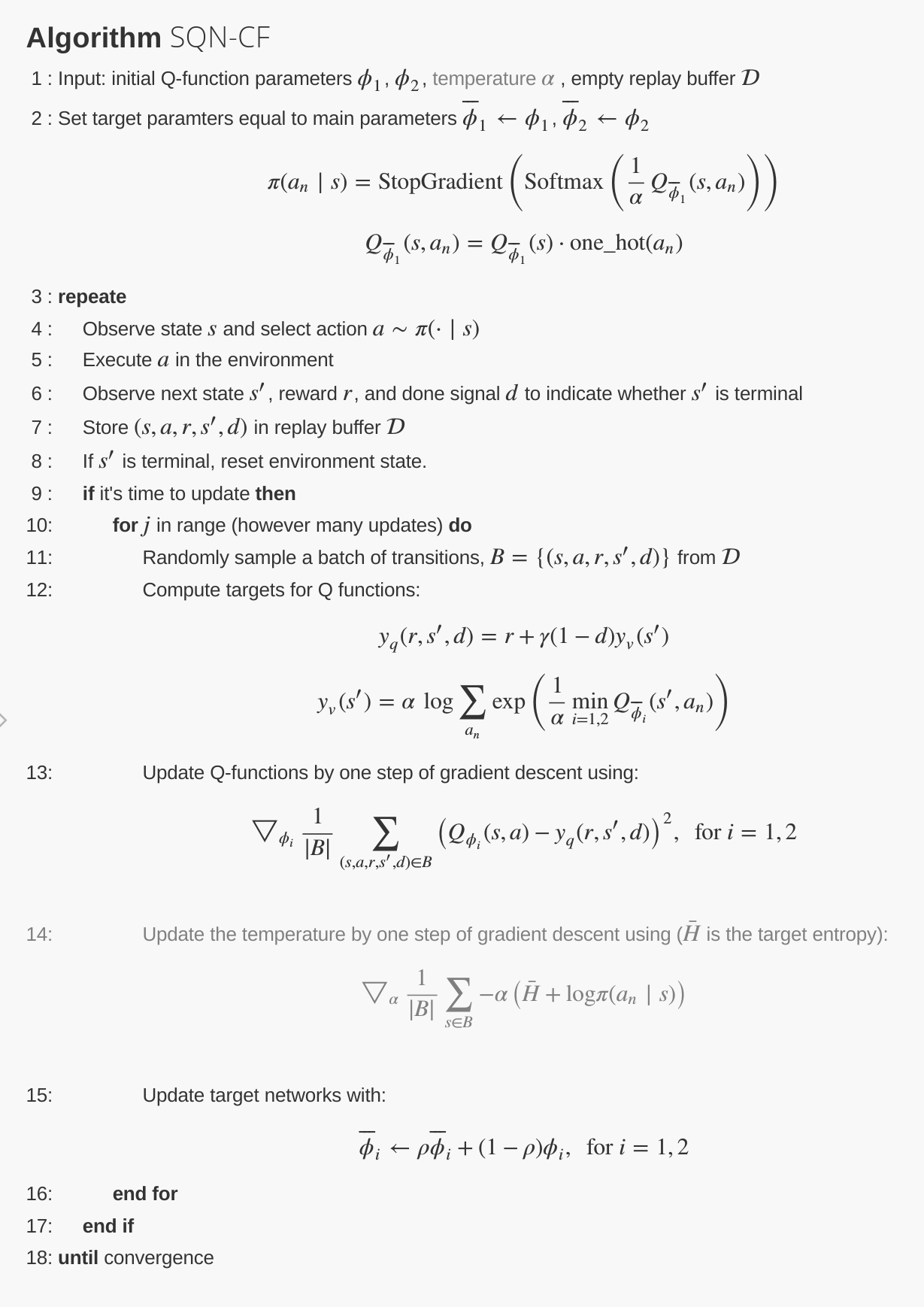}
    \caption{The SQN-CF algorithm. }
    \label{fig:sqn-cf2}
\end{figure}

The new soft Bellman backup in Eq.(\ref{eq:backup1}) can be further reduced:
\begin{align}
Q(\st, \at)
&= \reward(\st, \at) + \discount\E{\stp\sim\pdyn}{\E{\atp\sim\policynew}{Q^\policyold(\stp, \atp) - \alpha \log \policynew(\atp|\stp)}}  \notag \\
&= \reward(\st, \at) + \discount\E{\stp\sim\pdyn}{ \alpha \log \sum_{i=1}^n \exp\left( Q^\policyold (\stp,a_{t+1}^i) / \alpha \right) }  \notag \\
&= \reward(\st, \at) + \discount\E{\stp\sim\pdyn}{ \alpha \log Z^\policyold(\stp) } \notag \\
&= \reward(\st, \at) + \discount\E{\stp\sim\pdyn}{ V^\policyold(\stp) } .\label{eq:backup1a}
\end{align}
With the new soft Bellman backup, we can construct the SQN-CF algorithm. The pseudo-code of SQN-CF is presented in Fig.\ref{fig:sqn-cf2}. There are two main differences between SQN-CF and SQN, one is that the policy is defined according to the target Q function, the other is different formula of the soft Bellman backup. The deduction of SQN-CF is general. It holds for SQL and we have SQL-CF. As for the continuous case of SQL-CF, SAC-CF can be derived.

\subsection{Q On Policy}
As pointed out in \cite{schulman2017equivalence}, SQL equals Soft Policy Gradient (SPG). With SQL-CF, the proof of the equivalence of SQL and SPG is straightforward. We can obtain the equality $Q_\theta(\st, \at) = V_\theta(\st) + \alpha \log \policy_\theta(\at|\st) $ with Eq.(\ref{eq:pi_new1}). By substituting $Q_\theta(\st, \at)$ with $V_\theta(\st)$ and $\policy_\theta(\at|\st)$, we have:
\begin{align}
&\bigtriangledown_\theta \E{\st,\at\sim\policy} {\frac{1}{2}||Q_\theta(\st, \at)-\Hat{Q}_t||^2}|_{\policy=\policy_\theta} \notag\\
&= \E{\st,\at\sim\policy} {(\bigtriangledown_\theta Q_\theta(\st, \at)) (Q_\theta(\st, \at)-\Hat{Q}_t) }|_{\policy=\policy_\theta} \notag \\
&= \E{\st,\at\sim\policy} {(\bigtriangledown_\theta V_\theta(\st) + \alpha \bigtriangledown_\theta \log \policy_\theta(\at|\st)) (V_\theta(\st) + \alpha \log \policy_\theta(\at|\st)-\Hat{Q}_t) }|_{\policy=\policy_\theta} \notag \\
&= \E{\st,\at\sim\policy} {\bigtriangledown_\theta V_\theta(\st)(V_\theta(\st) -\Hat{Q}_t)  + \alpha \bigtriangledown_\theta \log \policy_\theta(\at|\st) (V_\theta(\st) + \alpha \log \policy_\theta(\at|\st)-\Hat{Q}_t) }|_{\policy=\policy_\theta} \notag \\
&= \E{\st,\at\sim\policy} {\bigtriangledown_\theta \frac{1}{2}||V_\theta(\st)-\Hat{Q}_t||^2
- \alpha \bigtriangledown_\theta \log \policy_\theta(\at|\st) ( \Hat{Q}_t - V_\theta(\st) - \alpha \log \policy_\theta(\at|\st)) }
|_{\policy=\policy_\theta}
\end{align}
The above equation is exactly the objective gradient of SPG \cite{SPG}. The first term is for Value estimation and the second term is for policy gradient.

In Eq.(\ref{eq:backup1}), the target is defined by one-step backup. We can extend it into n-step backup and verify that Eq.(\ref{eq:backup2}) still holds, that is to say, we have the n-step SQL-CF algorithm, i.e. the QOP alogrithm. 
The equivalence of SQL and SPG indicates the on-policy nature of SQL. From SQL-CF, we know that the soft Bellman backup of SQL-CF provide a policy improvement in the form of $Q$ value ( similar to the policy improvement in the PG algorithm ), which underlies the on-policy nature of SQL.

\section{EXPERIMENT}\label{header-n4}

In the last section, we propose three algothrims: SQN, SQN-CF, QOP. SQN-CF is based on SQN and QOP is based on SQN-CF. All the three algorithm can be used to solve RL environments. The code of the three algorithms can be found in \url{https://github.com/createamind/DRL}. SQN is to DQN what SAC is to DDPG. DQN and DDPG are the discrete version and the continuous version of Q learning, respective. Similarly SQN and SAC are the discrete version and the continuous version of soft Q learning, respective. So it is reasonable to assume SQN can have all the advantages of SAC  comparing with DQN and DDPG.

In this work, we are particularly interested in the on-policy Q learning, which is seldom studied in the RL community before. In this section, we focus on QOP and show the effectiveness and efficiency of the QOP alogrithm. We apply QOP to train GRF agent. GRF is a novel open-source reinforcement learning environment where agents learn to play football \cite{GRF}. The project can be found in \url{https://github.com/google-research/football}. There are several representations available for the observation state. We choose the simple115 representation, i.e. 115 floats to represent one frame state of observation. The action space is composed of 19 discrete actions. The reward is sparse, denoting the score of a team.

\begin{figure}
    \centering
    \includegraphics[width=0.759\textwidth]{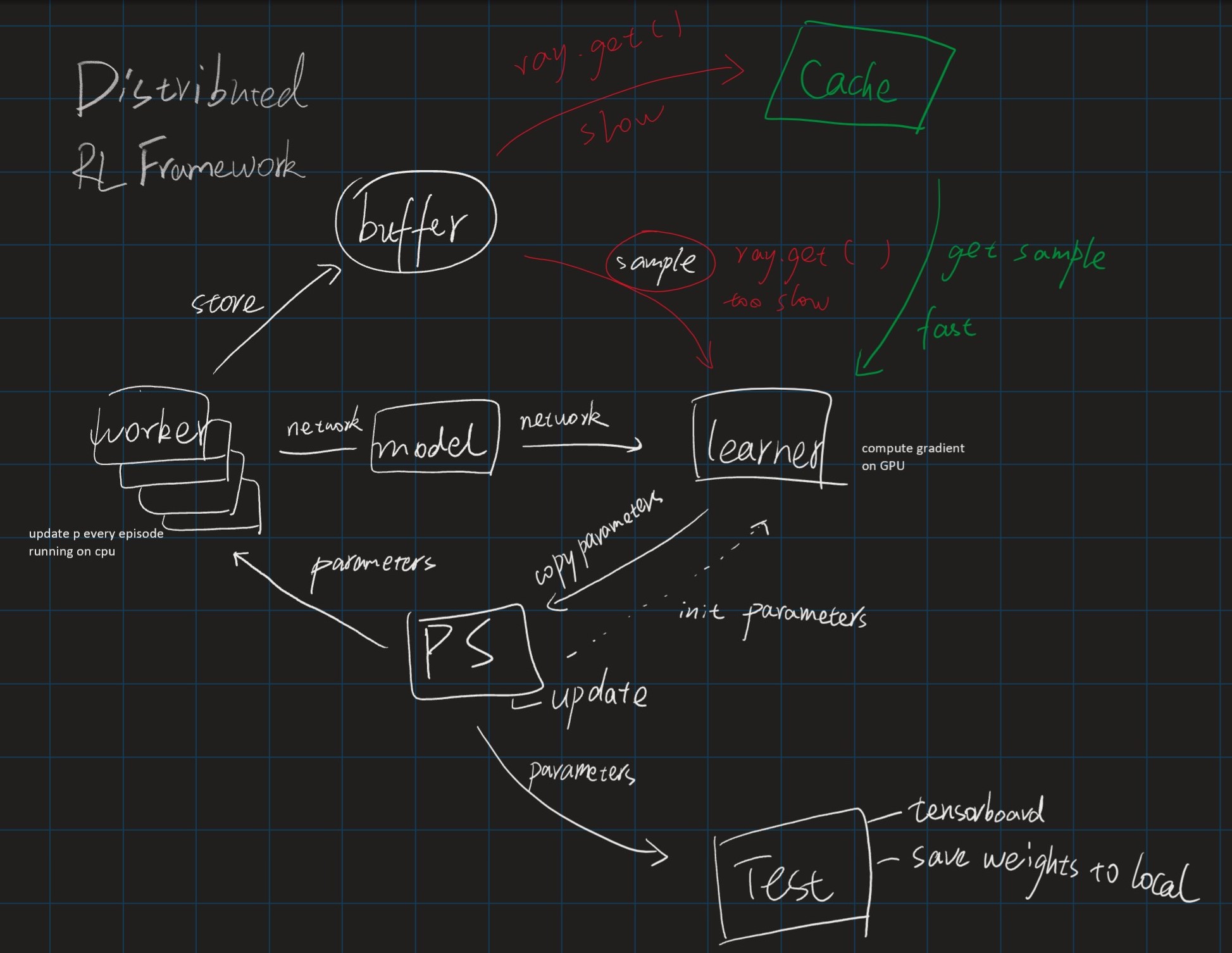}
    \caption{The training framework.}
    \label{fig:ddrlframework}
\end{figure}

\begin{figure}
    \centering
    \includegraphics[width=0.8\textwidth]{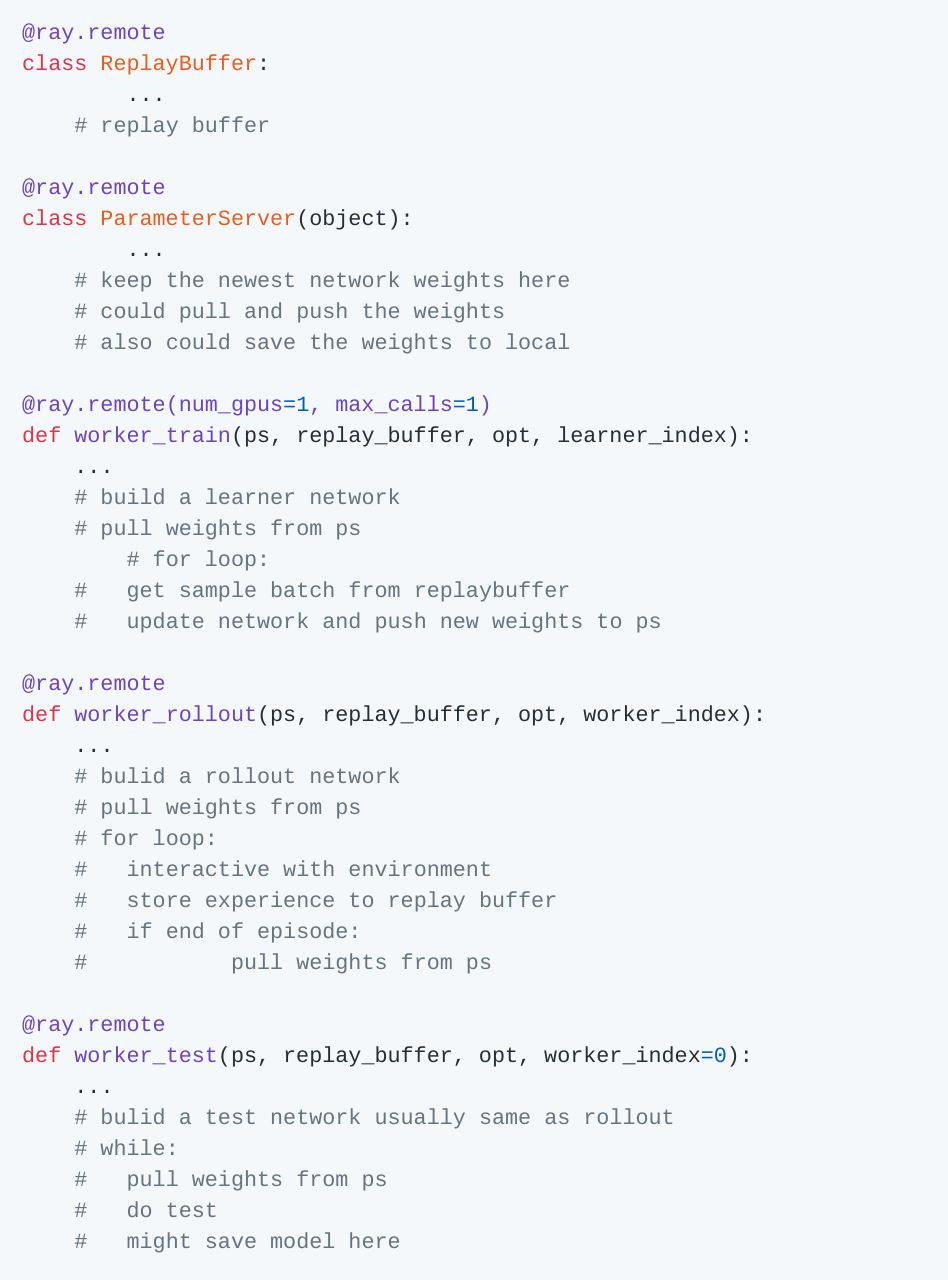}
    \caption{The APIs of our training framework.}
    \label{fig:ddrl5}
\end{figure}

We build a training framework based on a distributed framework called Ray \cite{Ray}. The project of Ray can be found in \url{https://github.com/ray-project/ray}.  The structure of our training framework is prsented in Fig. \ref{fig:ddrlframework}. The main components of our framework are two remote actors, i.e. ReplayBuffer, ParameterServer, and three romote functions, i.e. train worker, roll-out worker, test worker. We also introduce a Cache for train worker, in order to reduce the waiting time of the trainer. Their functions are presented in Fig. \ref{fig:ddrl5}.  The code of our training framework can be found in \url{https://github.com/createamind/Distributed-DRL}.

\begin{figure}
    \centering
    \includegraphics[width=0.759\textwidth]{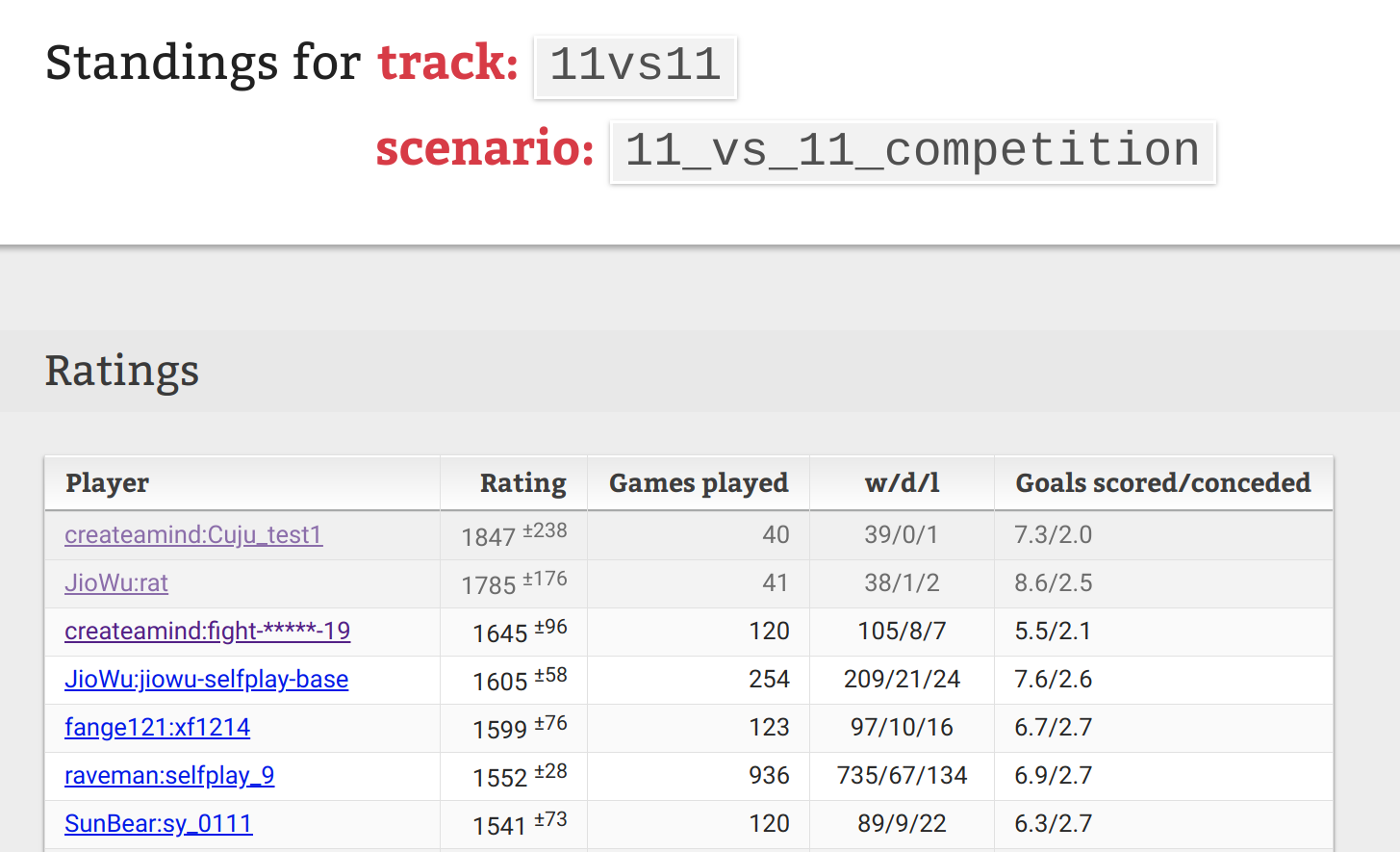}
    \caption{QOP can generate strong GRF agent, which stands in the first place of the GRF leaderboard.}
    \label{fig:leaderboard}
\end{figure}

Our agent is trained by 'self-play'. Training data is generated by current agent playing with diverse opponents. In training 80\% of the games are played against a target opponent, and 20\% play against past versions which are stored in a weights pool.
We did not adopt the pure self-play training because we thought that the external entropy reward breaks the zero-sum rule of self-play and can have negative influence in the training process. 
The target opponent is updated when current agent can win the target opponent with winning rate of 60\%. The Q network is represented as a (600, 800, 600) MLP with ReLU activation.
The QOP algorithm can be seen as n-step SQN-CF. Generalized Advantage Estimation (GAE) is the generalized version of the normal advantage estimation. GAE backup can reduce to n-step backup when $\lambda = 1$. In the experiment, we use n-step backup for the sake of simplicity but we think GAE backup is a better option. GAE backup is more stable and efficient than n-step backup. 
In the training process, we find that the reuse ratio of sampling data can have great impact on the training speed. We tune the reuse ratio by controlling the network update frequency. The smaller reuse ratio we take, the higher training speed we get, even when the reuse ratio is less than 1. It indicates the on-policy training nature of QOP. The QOP algorithm can generate strong GRF agent with diverse attack and counter-acttack strategies, and as shown in Fig. \ref{fig:leaderboard} our agent Cuju win the first place of the GRF leaderboard in April of 2020.

\section{CONCLUSION}\label{header-n5}

Deep Q Network is a simple algorithm,  yet it is surprisingly successful. Since its announcement many improvements have been made to enhance its performance, yet the inherent problem of DQN, i.e. the exploit-explore balance, remains. In this work, we investigate the effect of entropy regularization on DQN and propose SQN. SQN is the soft version of DQN, just like SAC (SDDPG) is the soft version of DDPG.  All the improvements on DQN, such as IMPALA\cite{espeholt2018impala}, APEX\cite{APEX}, R2D2\cite{R2D2} etc., can be equally effective on SQN.
For deep Q network, an exploratory factor like $\epsilon$-greed needs to be artificially added and tuned for training. While SQN can automatically balance exploit-exploit through the temperature $\alpha$. Compared with DQN, SQN handles exploit-exploit trade-off more elegantly and the performance is much better.
We show that SQL can have the corrective feedback effect and propose SQL-CF. The soft backup of SQL-CF can be seen as a policy improvement in the form of Q, and it underlies the equivalence of SQL and Soft Policy Gradient (SPG). With the insight of the on-plicy nature of SQL, we propose an on-policy version of deep Q learning algorithm, i.e. Q On-Policy (QOP). Different from n-step DQN, QOP has rigorous policy improvement guarantee. We experiment with QOP on GRF and the QOP algorithm shows great stability and efficiency in training.

\bibliographystyle{apalike}
\bibliography{ref.bib}

\end{document}